\documentclass[a4paper]{article}

\usepackage[english]{babel}
\usepackage[utf8x]{inputenc}
\usepackage[T1]{fontenc}

\usepackage[a4paper,top=3cm,bottom=2cm,left=3cm,right=3cm,marginparwidth=1.75cm]{geometry}

\usepackage{amsmath}
\usepackage{graphicx}
\usepackage[colorinlistoftodos]{todonotes}
\usepackage[colorlinks=true, allcolors=blue]{hyperref}

\title{Deep learning research landscape \& roadmap in a nutshell: past, present and future - Towards deep cortical learning}
\author{Aras R. Dargazany}

\begin{document}
\maketitle

\begin{abstract}
The past, present and future of deep learning is presented in this work. 
Given this landscape \& roadmap, we predict that deep cortical learning will be the convergence of deep learning \& cortical learning which builds an artificial cortical column ultimately.  
\end{abstract}

\section{Past: Deep learning inspirations}
\label{sec:past}
\begin{figure*}[ht!]
\centering
\includegraphics[width=\textwidth]{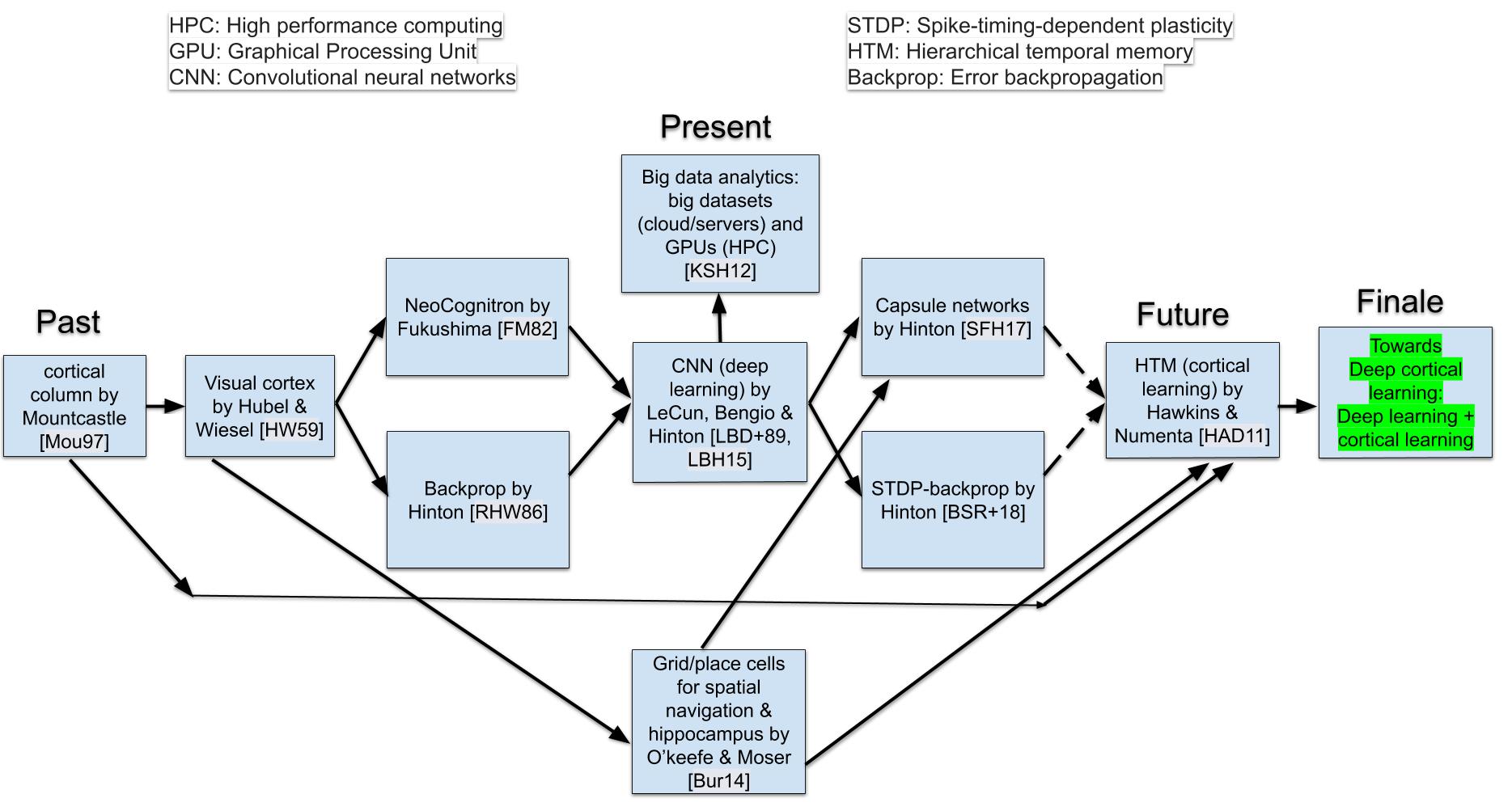}
\caption{Deep learning research landscape \& roadmap: past, present, future. The future is highlighted as deep cortical learning.}
\label{fig:my-visual-research-roadmap}
\end{figure*}

\begin{figure*}[ht!]
    \centering
    \includegraphics[width=\textwidth]{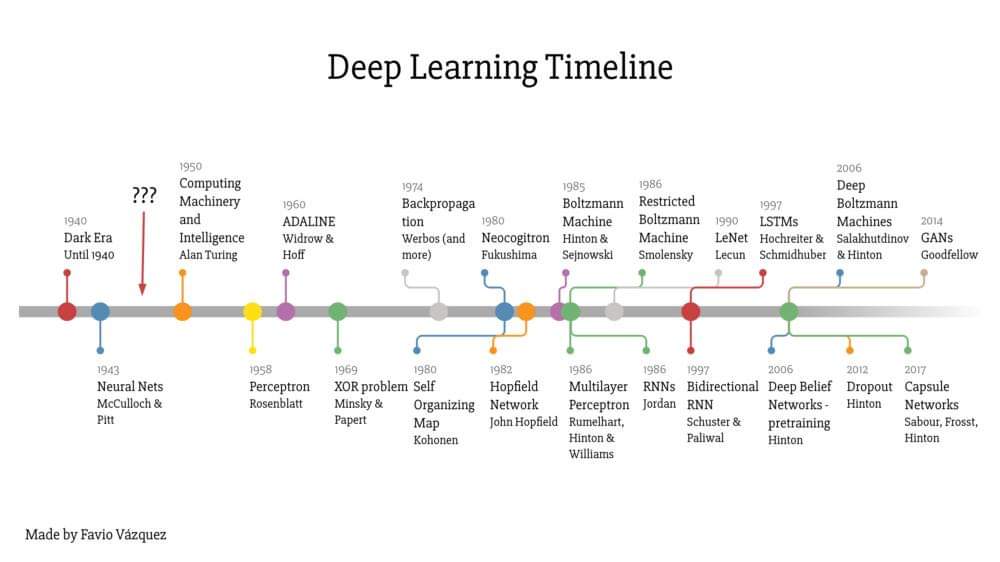}
    \caption{Neural nets origin, timeline \& history made by Favio Vazquez}
    \label{fig:dl-timeline}
\end{figure*}

Deep learning horizon, landscape and research roadmap in nutshell is presented in this figure \ref{fig:my-visual-research-roadmap}.
%
The historical development and timeline of deep learning \& neural network is separately illustrated in figure \ref{fig:dl-timeline}.
The Origin of neural nets \cite{wang2017origin} is thoroughly reviewed in terms of the evolutionary history of deep learning models.
%
Vernon Mountcastle discovery of cortical columns in somatosensory cortex~\cite{mountcastle1997columnar} was a breakthrough in brain science.
%
The big bang was the discovery of Hubel \& Wiesel of simple cells and complex cell in visual cortex \cite{hubel1959receptive} which won the Nobel prize for this discovery in 1981.
This work was heavily founded on Vernon Mountcastle discovery of cortical columns in somatosensory cortex~\cite{mountcastle1997columnar}.
%
After the discovery of Hubel \& Wiesel, Fukushima proposed a pattern recognition architecture based on the simple cell and complex cell discovery, known as NeoCognitron \cite{fukushima1982neocognitron}.
In this work, a deep neural network was proposed using simple cell layer and complex cell layer repeatedly.
%
In 80s and maybe a bit earlier backpropagation have been proposed by multiple people but the first time it was well-explained and applied for learning neural nets was done by Hinton and his colleagues in 1987 \cite{rumelhart1986learning_2}.

\section{Present: Deep learning by LeCun, Bengio and Hinton}
Convolutional nets was invented by LeCun \cite{lecun1989backpropagation} which led to
deep learning conspiracy which also started by the three founding fathers of the field: LeCun, Bengio and Hinton \cite{lecun2015deep}.
The main hype in deep learning happened in 2012 when the state-of-the-art result in Imagenet classification and TIMIT speech recognition task were dramatically reduced using an end-to-end deep convolutional network~\cite{krizhevsky2012imagenet} and deep belief net~\cite{hinton2012deep}. 

The power of deep learning is scalability and the ability to learn in an end-to-end fashion.
In this sense, deep learning architectures are capable of learning big datasets such as Imagenet \cite{krizhevsky2012imagenet,goyal2017accurate} and TIMIT using multiple GPUs in an end-to-end fashion meaning directly from raw inputs, all the way the desired outputs.
Alexnet \cite{krizhevsky2012imagenet} used two GPUs for Imagenet classification which is a very big dataset of images, almost 1.5 million images of size 215x215.
Kaiming He et al. \cite{goyal2017accurate} proposed a highly scalable approach for training on Image using 256 GPUs for almost an hour which shows an amazingly powerful approach based stochastic gradient descent for applying big cluster of GPUs on huge datasets.
Very many application domains have been revolutionized using deep learning architectures such as  image classifications \cite{krizhevsky2012imagenet}, machine translation \cite{wu2016google,johnson2016google}, speech recognition \cite{hinton2012deep}, and robotics \cite{mnih2015human}.

The Nobel Prize in Physiology or Medicine 2014 was given to John O’Keefe, May-Britt Moser and Edvard I. Moser “for their discoveries of cells that constitute a positioning system in the brain.”~\cite{burgess20142014}. 
This study of cognitive neuroscience shed light on how the world is represented within the brain.
Hinton's Capsule network~\cite{sabour2017dynamic} and Hawkins' cortical learning algorithm~\cite{hawkins2011hierarchical} are highly inspired by this Nobel-prize winning work~\cite{burgess20142014}.

\section{Future: Brain-plausible deep learning \& cortical learning algorithms}
The main direction and inclination in the deep learning for future is the ability to bridge the gap between the cortical architecture and deep learning architectures, specifically convolutional nets.
In this quest, Hinton proposed capsule network \cite{sabour2017dynamic} as an effort to get rid of pooling layers and replace it with capsules which are highly inspired bu cortical mini-columns in cortical columns and layers and include the location information or pose information of parts.

Another important quest in deep learning is understanding the biological root of learning in our brain, specifically in our cortex.
Backpropagation is not biologically inspired and plausible. 
Hinton and the other founding fathers of deep learning have been trying to understand how backprop might be feasible biologically in brain.
Feedback alignment~\cite{lillicrap2016random} and spike time-dependent plasticity or STDP-based backprop~\cite{bartunov2018assessing} are some of the works which have been done by Timothy Lillicrap, Blake Richards, and Hinton in order to model backprop biologically based on the pyramidal neuron in the cortex.

In the far future, the main goal should be the merge of two very independent quest to build cortical structure in our brain:
The first one is heavily target by the big and active deep learning community;
The second one is targeted independently and neuroscientifically by Numenta and Geoff Hawkins \cite{hawkins2011hierarchical}.
These people argue that the cortical structure and our neocortex is the main source of our intelligence and for building a true intelligent machine, we should be able to reconstruct the cortex and to do so, we should first focus more on the cortex and understand what cortex is made out of.

\section{Finale: Deep cortical learning as the merge of deep learning and cortical learning}
By merging deep learning and cortical learning, a very more focused and detailed architectures, named deep cortical learning might be created. 
We might be able to understand and reconstruct the cortical structure with much more accuracy and have a better idea what the true intelligence is and how artificial general intelligence or AGI might be reproducible.
Deep cortical learning might be the algorithm behind one cortical column in the neocortex.

\newcommand{\etalchar}[1]{$^{#1}$}

\end{document}